\documentclass[journal]{IEEEtran}

\ifCLASSINFOpdf
\else
   \usepackage[dvips]{graphicx}
\fi

\usepackage{url}
\usepackage{cite}
\usepackage{hyperref}
\usepackage{amsmath,amssymb}
\usepackage{multirow,multicol}
\usepackage{xcolor}

\DeclareMathOperator{\EX}{\mathbb{E}}

\usepackage{graphicx}

\begin{document}

\title{Metric-Based Few-Shot Learning\\for Video Action Recognition}

\author{
Chris~Careaga$^1$, Brian~Hutchinson$^{1,2}$, Nathan~Hodas$^2$ and~Lawrence~Phillips$^2$\\
    $^1$Computer Science Department, Western Washington University, Bellingham, WA, USA\\
$^2$Data Science and Analytics, Pacific Northwest National Laboratory, Richland, WA, USA}

\maketitle

\begin{abstract}
In the few-shot scenario, a learner must effectively generalize to unseen
classes given a small support set of labeled examples.  While a relatively
large amount of research has gone into few-shot learning for image
classification, little work has been done on few-shot video classification.  In
this work, we address the task of few-shot video action recognition with a set
of two-stream models.  We evaluate the performance of a set of convolutional
and recurrent neural network video encoder architectures used in conjunction
with three popular metric-based few-shot algorithms.  We train and evaluate
using a few-shot split of the Kinetics 600 dataset.  Our experiments confirm the
importance of the two-stream setup, and find prototypical networks and  pooled
long short-term memory network embeddings to give the best performance as
few-shot method and video encoder, respectively.  For a 5-shot 5-way task, this
setup obtains 84.2\% accuracy on the test set and 59.4\% on a special
``challenge'' test set, composed of highly confusable classes.
\end{abstract}

\begin{IEEEkeywords}
few-shot, video classification, action recognition, convolutional neural network, lstm
\end{IEEEkeywords}

\IEEEpeerreviewmaketitle

\section{Introduction}

\IEEEPARstart{D}{eep} learning has shown impressive successes across many
domains, from spoken 
and
natural language 
\cite{dahl2012context,hinton2012deep,chan2016listen} 
\cite{sutskever2014sequence,vaswani2017attention} 
to computer
vision \cite{he2016deep, krizhevsky2012imagenet} to bioinformatics
\cite{evans2018denovo}.  Unfortunately, their success often depends on having
access to large amounts of labeled training data, of which collecting and labeling
is very time-consuming and costly.  Furthermore, for many applications the
classes of interest belong to a long tail, where it is impractical to obtain a
large set of labeled instances for most classes.  This limitation motivates the
scenario we are interested in solving: training a deep model using very few
labeled examples. This setting is known as {\it few-shot} learning. 

In recent years, many researchers have introduced few-shot learning methods,
mainly for the task of image classification.  Most methods are based on the
idea of {\it meta-learning} (i.e. learning to learn).  One popular approach to
few-shot meta-learning can be described as {\it metric-based}
\cite{koch2015siamese, vinyals2016matching, snell2017prototypical,
sung2018learning}.  In these methods, data points are embedded into a feature
space (e.g. via a neural network) and then predictions are made using the
distances between the embeddings of query images and those of a small {\it support} set of
labeled training points.  Classification loss can be back-propagated to the
embedding network over many small tasks (``episodes'') in order to create an
embedding function that is effective across a range of distinct classes.

While relatively little attention has been given to the problem of few-shot video
classification, deep learning approaches to video classification have been
studied extensively over the past decade \cite{karpathy2014large,
simonyan2014two, donahue2015long, tran2015learning, yao2015describing,
feichtenhofer2016convolutional, wang2016temporal, girdhar2017action,
carreira2017quo}.  Most often, video classification is accomplished by first
extracting frame-level features of a video using convolutional neural networks
(CNN), and then aggregating the features over time to yield a video-level
representation. Many variants of both the frame-level feature extractor and the
aggregation technique have been developed.  The video encoders we present in
this paper draw from the findings in this body of traditional video
classification work. Rather than optimizing for direct classification, we
instead aim to produce discriminative embeddings for few-shot classification.

Recently, researchers have begun to address the task of few-shot
learning for video action recognition. Thanker and Krishnakumar
\cite{thakerkshot} made use of dynamic images and memory augmented neural
networks (MANN) \cite{santoro2016meta} to perform few-shot action recognition
on the Kinetics 400 dataset \cite{carreira2017quo}.  Zhu and Yang
\cite{zhu2018compound} propose another MANN based architecture they call the
``compound memory network'' (CMN). Their method, obtains 78.9\% accuracy on a
5-way 5-shot task on the Kinetics 400 \cite{carreira2017quo} dataset. Bishay et
al.  propose a video-specific version of the Relation Network
\cite{sung2018learning} and leverage a large 3D CNN backbone network
to generate embeddings.  They manage to out-perform the CMN,  but only when
using a larger backbone.  Hu et. al \cite{hu2019weakly} propose a method of
explicitly modeling the composition of semantic features for few-shot
recognition. They achieve an 83.1\% accuracy for a 5-shot 5-way task on the
Kinetics 400. Cao et. al \cite{cao2019few} propose a temporal alignment method in order to better
model the temporal dynamics of video. They achieve a state-of-the-art 85.8\%
accuracy on the Kinetics 400 using a ResNet50 \cite{he2016deep} backbone. Daesik et al.
\cite{daesik2017matching} propose a variation of an existing few-shot method
called Matching Networks \cite{vinyals2016matching}, with the addition of a
memory-based embedding.  Zou et al.  \cite{zou2018heirarchical} propose a
complex memory-based model. Both methods evaluate on cleaner, small datasets
\cite{liu2009recognizing, rodriguez2008action, kuehne2011hmdb}. 

In this work, we introduce a set of approaches to few-shot video action
recognition. We propose and evaluate several video encoder architectures across
three few-shot methods.  We also investigate the importance of including an
optical flow feature stream.  We train and evaluate our models using a 
``few-shot'' split of the Kinetics 600 \cite{carreira2017quo, carreira2018a}
dataset.  In addition to training and validation sets, the splits include one
general test set (with randomly drawn classes) and one ``challenge'' test set,
whose classes are highly similar (all aggressive/violent actions - e.g.
kicking, boxing).  
Prior work in few-shot video action
recognition has leveraged complex methods and large backbone CNNs. We find
video encoder architectures that allow ``simple'' few-shot methods (e.g.
Prototypical Networks \cite{snell2017prototypical}) to yield performance
comparable to state-of-the-art models, while utilizing far fewer parameters.

\section{Few-shot Background}
In the few-shot setting, rather than aiming to generalize to previously unseen
examples, one aims to generalize to previously \textit{unseen classes}, given
only a small set (e.g.  1, 5) of labeled examples per class.  Often few-shot
learning is posed as an instance of {\it meta-learning} (i.e. learning to
learn).  Formally, let $E$ denote a set of $n$ classes. Assume inputs $x$
come from some set ${\cal X}$.  Let $S_{E}$ be a ``support'' set, containing
$k$ input-label pairs for each of the $n$ classes in $E$.  We seek a meta-model
$M_\theta$, parameterized by $\theta$, that takes any support set $S_{E}$ and
produces a new classifier $C_{S_{E},\theta} : {\cal X} \to E$.  That is, with
only the small amount of data contained in support set $S_{E}$, the meta model
$M_\theta$ allows us to construct a classifier capable of classifying novel $x
\in {\cal X}$ as one of the classes in $E$.

The challenge of meta-learning lies in training the meta-model; namely, in
estimating its parameters, $\theta$. In order to match the evaluation
environment, meta-models are often trained using {\it episodic} training
\cite{vinyals2016matching}. In each episode during episodic training, a set of
classes $E$ is sampled from a larger set of classes called meta-train.  Given
$E$, a support set $S_{E}$ with $k$ input-output pairs per class in $E$ is
sampled.  A disjoint {\it query set} $Q_{E}$ is also sampled, with distinct
input-output pairs drawn from the same set of classes $E$. Together, $S_{E}$
and $Q_{E}$ are called an {\it episode}.  Meta-model $M_\theta$ takes $S_{E}$
and produces $C_{S_E,\theta}$, which is used to make predictions on the query
set $Q_{E}$.  The loss on the query set is back-propagated to adjust the
meta-model parameters, $\theta$. Formally, training the meta-model involves
solving the following problem, where we drop the subscripts to simplify notation:

\begin{equation} 
    \arg\min_{\theta} \EX_{E \sim \mathcal{T}} \left( \EX_{Q,S \sim E} \left( \sum\limits_{(x,y) \in Q} L(C_{S,\theta}(x), y) \right) \right)
\end{equation}

Here $L$ denotes cross-entropy loss.  Once trained, the meta-model can be
evaluated by averaging classification accuracy over many  episodes drawn from a
set of unseen classes called meta-test.

In this paper, we focus on metric-based few-shot methods. This means
the parameters of the meta-learner belong to the encoder network that
embeds videos into a metric space. The model $C_{S_E,\theta}$ produced by
$M_\theta$ may itself be non-parametric; for example, making its decision only
by distances between the embedding of a novel query video $x$ and the
embeddings of the support set videos. If successful, the meta-learner defines
a general embedding space that is agnostic to the particular classes ($E$)
provided; it embeds examples to capture notions of similarity spanning all
classes within the video action recognition domain. 

\section{Methods}

\begin{figure}
    \centering
    \includegraphics[width = 250pt]{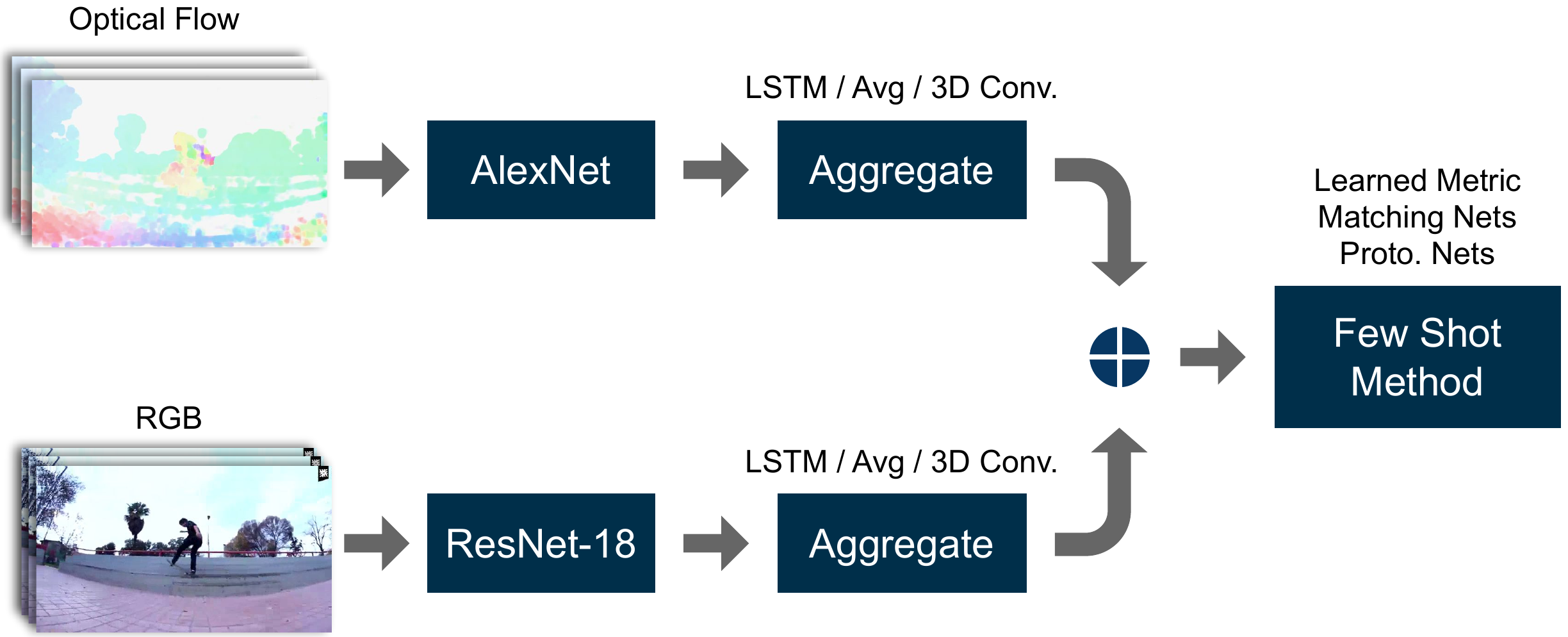}
    \caption{The general structure of our proposed models. Frame-level features
    are extracted and aggregated across time to create a fixed-length
    representation for each stream. The streams are concatenated
    and fed into the few-shot method.}
    \label{fig:system_diagram}
\end{figure}

Our few-shot architecture structure is illustrated in
Fig.~\ref{fig:system_diagram}.  First, CNNs are applied frame-wise
to the RGB and optical flow streams. The resulting (flat or spatial) streams
are aggregated into a fixed length representation per video, which are then fed
into a few-shot algorithm. Details are provided below. 

\subsection{Flow Feature Extraction} 
While all of the information in the video is contained in the sequence of RGB
frames, $x_{1:T}^{rgb}$ (where $T$ is the number of frames), researchers have
shown that deep learning models perform significantly better at action
recognition when explicit optical flow features are extracted and fed into the
model \cite{feichtenhofer2016convolutional, simonyan2014two}. In the few-shot
scenario, where data is by definition limited, we seek the richest video
representation possible. We therefore
extract an optical flow feature stream, denoted $x_{1:T}^{flw}$ and feed this
in to our models, as described below.  

\subsection{Frame-level Encoders}
For the RGB and flow feature streams, we use two popular pre-trained CNN
architectures to extract higher level frame representations, respectively:
ResNet-18 \cite{he2016deep} and AlexNet \cite{krizhevsky2012imagenet}.  Both
are distributed with the PyTorch \cite{paszke2017automatic} library, and are
pre-trained on ImageNet \cite{deng2009imagenet}. We find the pre-trained
weights provide a good initialization for both RGB and optical flow.  
Both architectures were selected to maintain a light memory footprint, with the 
space utilization balance tilted in favor of the more informative feature stream, RGB.
Both models are fine-tuned during meta-training.

\subsubsection{ResNet-18}
The ResNet-18 model is used to encode each RGB frame, $x^{rgb}_{i}$. To
represent each frame we extract $512$-dimensional features immediately before the
ResNet-18's final linear layer. For our third few-shot method (learned distance metric), the encoder's output is fed into a CNN, and thus must have spatial dimensions.
In this case, feature maps
are extracted from the ResNet before the last residual block.  This provides
$14 \times 14$ feature maps which we further process using a $3 \times 3$
convolution, with a stride of 2, followed by $2 \times 2$ average pooling with a
stride of $2$.  This generates $256$ $6\times 6$ feature maps.
The RGB features extracted from the ResNet for the $i$th frame are denoted
$\hat{x}^{rgb}_{i}$.

\subsubsection{AlexNet}
The AlexNet model is used to encode each optical flow frame, $x^{flw}_{i}$.
Like the ResNet, we remove the final linear layers, and the AlexNet produces
spatial features ($256$ $6\times 6$ feature maps). When flat embeddings are needed, the feature maps
are flattened to yield a $9216$-dimensional vector and passed through a linear layer 
to produce a $512$-dimensional embedding. The features generated by the
AlexNet for the $i$th frame are denoted $\hat{x}^{flw}_{i}$.

\subsection{Aggregation Techniques}
The embeddings generated by the ResNet-18 and AlexNet need to be aggregated
over time to produce a fixed-length embedding, $\hat{x}$, for each video. We
consider four aggregation strategies.

\subsubsection{Pooling}
We simply average the vectors across time.\footnote{In preliminary
experiments we found average pooling to outperform max pooling, so we use only
the former in this work.}
\begin{equation}
\hat{x} = \text{Pool}(\hat{x}^{rgb}_{1:T}) : \text{Pool}(\hat{x}^{flw}_{1:T})
\end{equation}

Here the Pool operation refers to average pooling and ``:'' denotes concatenation.

\subsubsection{LSTM}
To capture the temporal dynamics of each video, it is intuitive to utilize a
recurrent model. Here we use the Long Short-Term Memory (LSTM) Network
\cite{hochreiter1997long} to yield a temporally-aware video-level
representation.

\begin{equation}
    \hat{x} = 
    \text{Pool}(\text{LSTM}(\hat{x}^{rgb}_{1:T})) : 
    \text{Pool}(\text{LSTM}(\hat{x}^{flw}_{1:T}))
\end{equation}

The LSTM operation consists of a single layer LSTM network with a hidden
representation size of $512$.  We find that averaging the LSTM hidden
representation over all timesteps yields faster training and better results.

\subsubsection{ConvLSTM}
When using the learned distance metric method, a vanilla LSTM cannot be used as
it expects flat features.  We solve this by using a Convolutional LSTM
(ConvLSTM) \cite{shi2015conv}. The linear sublayers of the LSTM are replaced by
convolutional layers.

\subsubsection{3D Convolutions}
Rather than combining flat embeddings, the spatio-temporal aspect of each video
can be captured via 3D convolutions. We propose to combine spatial
feature maps of each frame over time as follows:

\begin{equation}
    \hat{x} = 
        \text{Pool}(\text{Conv3d}(\hat{x}^{rgb}_{1:T})) : 
        \text{Pool}(\text{Conv3d}(\hat{x}^{flw}_{1:T}))
\end{equation}

Where the Conv3d operation consists of two 3D convolutional layers with a
kernel size of $3 \times 3 \times 3$ and a stride of $1 \times 1 \times 1$, with a ReLU activation in-between.
If flat features are required,
then adaptive average pooling is applied to create a $512$-dimensional vector. If
spatial features are required, adaptive average pooling is applied to create
$512$ $6\times 6$ feature maps.

\subsection{Few-shot Methods}
Once video-level embeddings are generated, various few-shot methods can be used
to classify each video using only the support set. We consider three popular
metric-based few-shot methods. 

\subsubsection{Matching Networks}
Matching Networks \cite{vinyals2016matching} are a few-shot method consisting
of an image embedding function parameterized by a CNN, along with a
metric-based classification in the embedding space. This method works by
embedding both the query set and support set. To classify an example from the
query set, the distance to each instance in the support set is calculated.  For
a given query input, $x$, classification scores are calculated by averaging the
negative squared euclidean distance from a query embedding to the support set
embeddings for each class:

\begin{equation}
z_k(x) = \frac{1}{|S_k|}\sum\limits_{(x_i, y_i) \in S_k} -d(f_{\theta}(x), f_{\theta}(x_i))
\end{equation}

The scores are then softmaxed to yield a probability distribution over possible
classes. For simplicity, we did not incorporate fully-contextual
embeddings in our implementation.

\subsubsection{Prototypical Networks} 
Prototypical Networks \cite{snell2017prototypical} are a few-shot method
similar to Matching Nets (equivalent in the one-shot case). Instead of
computing the distance from an embedded query to each example in the support
set, a single embedding (``prototype'') is created to represent each class. The
prototype, $c_k$, is calculated by averaging support set embeddings within each class.

\begin{equation}
c_k = \frac{1}{|S_k|} \sum\limits_{(x_i, y_i) \in S_k} f_{\theta}(x_i)
\end{equation}

The negative distances to each prototype are then used as classification logits:

\begin{equation}
z_k(x) = -d(f_{\theta}(x), c_k)
\end{equation}

\subsubsection{Learned distance metric}
We also propose a variant of the Prototypical Nets that is inspired by the 
Relation Network \cite{sung2018learning} few-shot method. This method
works by computing prototypes for each class, but instead of using a fixed
distance metric to compute classification scores, a small CNN, $g$, with
parameters $\phi$ is used to calculate the ``distance'' between a query and
each of the class prototypes:

\begin{equation}
z_k(x) = g_{\phi}(f_{\theta}(x) : c_k)
\end{equation}

The CNN $g$, consists of two convolutional layers, each with a $3\times 3$ kernel with a
stride of $1$, and each followed by ReLU.  The feature maps are then average
pooled, flattened, mapped to a scalar score via a linear layer.

\section{Experiments}

\subsection{Experimental Setup}

\subsubsection{Kinetics 600}
Kinetics 600 \cite{carreira2017quo, carreira2018a} is a collection of YouTube
videos categorized by action. The traditional splits are not suitable for the
few-shot scenario, so we propose custom splits for training, validation and
testing. We design our splits to mirror the structure of popular image few-shot
benchmark dataset, \textit{miniImageNet} \cite{ravi2017optimization}.  The
training and validation sets consist of 64 and 16 classes, respectively. We
define two test sets to evaluate the effectiveness of models in different
conditions. The first test set consists of 20 randomly sampled classes as is
typically used to quantify the generalization of the model to unseen classes.
The second, a ``challenge'' test set, consists of closely related classes (all
aggressive actions, e.g.  kicking, punching), and is used to determine the
model's ability to discriminate between very similar classes. Using Kinetics
600 instead of Kinetics 400 (e.g. as used in \cite{zhu2018compound}) offers two
key advantages: 1) like miniImageNet, each class has at least 600 instances
(the splits in \cite{zhu2018compound} are limited to 100 instances per class),
and 2) the larger number of classes facilitates the creation of our
``challenge'' test set.

\subsubsection{Preprocessing} \label{ssec:exptspreproc}
For each video, all RGB frames are extracted and downsampled to one frame
per second.  To be consistent with the expected inputs of the pre-trained
ResNet-18, the frames are resized to $224 \times 224$ and the RGB
features are standardized using the same mean and standard deviation that is
used by the pre-trained ResNet-18.  Pairs of frames are also sampled from each
video in order to compute optical flow features.  The frame pairs are resized
to $224 \times 224$ and  optical flow is computed using the F\"{a}rneback
algorithm \cite{farneback2003two} implemented in the OpenCV library
\cite{bradski2000opencv}. The flow is thresholded into the range $[-20,20]$,
rescaled into the range $[0, 1]$, and then standardized.  The pre-trained
AlexNet expects a 3-channel input, but the optical flow only has two,
(vertical and horizontal components), so we pad a zero third channel before 
feeding into the AlexNet. To ensure each video has a sufficient
number of frames, very short videos (fewer than five frames) are discarded from
the episode.

\subsubsection{Few-shot Setup}
All experiments are $n=5$~way $k=5$~shot; that is, meta-training episodes
consist of five examples from five classes for the support set and five
examples from the same five classes for the query set. We train for 25,000
episodes, checking an early stopping criterion on the validation set every 500
episodes.  
Very little hyperparameter tuning was performed. 
We use a fixed learning rate of $10^{-5}$ and
optimize using the Adam optimizer \cite{kingma2015adam}.  After training, test
results are obtained by randomly selecting and evaluating on 1000 episodes from
the test set, each with 10 queries per class. 

\subsection{Results and Analysis}

We first evaluate combinations of four aggregation strategies and three
few-shot methods; these results are reported in
Table~\ref{tab:compare_enc_and_method}. The most notable trend is the gap
between the general and challenge test sets, with the general test set
obtaining $\sim25$\% higher accuracy.  This suggests that in many realistic use
cases, where the distinction between classes is fine-grained, there is still
room for significant improvement in video few-shot learning.  Another clear
trend is the consistently superior performance of Prototypical Networks;
regardless of the aggregation method, Prototypical Networks give the highest
accuracy.  Comparing aggregation strategies, we observe that the LSTM gives the
highest performance, although simple averaging is a very competitive alternative.

\begin{table}
    \renewcommand{\arraystretch}{1.3}
    \caption{Test set accuracy and 95\% confidence interval, contrasting aggregation approaches (columns) and few-shot methods (rows), averaged over 1000 test episodes.}
    \label{tab:compare_enc_and_method}
    \centering
    \begin{tabular}{|l|c|c|c|c|}\hline
        \multicolumn{5}{|c|}{\bf General Test Set}\\\hline
        {\bf Method} & {\bf Averaging} & {\bf LSTM} & {\bf ConvLSTM} & {\bf 3dConv}\\\hline
        Prototypical & 83.5 $\pm$ 0.46 & {\bf 84.2} $\pm$ 0.44 & 77.9 $\pm$ 0.53 & 78.8 $\pm$ 0.51 \\\hline
        Matching & 79.1 $\pm$ 0.55 & 81.1 $\pm$ 0.50 & 75.7 $\pm$ 0.56 & 75.7 $\pm$ 0.56\\\hline
        Learned  & 77.9 $\pm$ 0.51 & - & 78.1 $\pm$ 0.51 & 74.1 $\pm$ 0.55\\\hline
        \multicolumn{5}{|c|}{\bf Challenge Test Set}\\\hline
        {\bf Method} & {\bf Averaging} & {\bf LSTM} & {\bf ConvLSTM} & {\bf 3dConv}\\\hline
        Prototypical & 58.5 $\pm$ 0.58 & {\bf 59.4} $\pm$ 0.59 & 53.6 $\pm$ 0.60 & 54.4 $\pm$ 0.60\\\hline
        Matching & 52.3 $\pm$ 0.57 & 54.6 $\pm$ 0.60 & 51.0 $\pm$ 0.60 & 49.2 $\pm$ 0.58\\\hline
        Learned & 51.5 $\pm$ 0.61 & - & 52.3 $\pm$ 0.61 & 50.3 $\pm$ 0.61\\\hline
    \end{tabular}
\end{table}

Table~\ref{tab:feature_ablation} lists the results of an ablation study,
designed to assess the importance of input features when using our best
performing model (LSTM aggregation paired with Prototypical Networks). First,
we compare the contributions of the RGB and flow streams. Using only the RGB
stream without flow, there is a small decrease in performance ($\sim1.5$\%).
In contrast, using only flow without RGB yields a significant drop in
performance ($\sim19.6\%$).  Next, we compare the effect of framerate.  We
report a ``single frame'' result, in which RGB and flow for a single randomly selected frame is
used; the performance is significantly worse than the using 1~fps RGB and flow
($\sim10$\%).  This suggests that a naive approach of reducing the video
few-shot task to an image few-shot task is ineffective.  We then
consider the effect of increasing either RGB and/or flow framerates to 2~fps,
and find that it yields little-to-no improvement in performance.  
Further increasing the framerate provided no improvements 
(not shown in table for brevity).

\begin{table}
\centering
\renewcommand{\arraystretch}{1.3}
\caption{Test set accuracy and 95\% confidence interval, varying input streams, averaged over 1000 test episodes.}
\label{tab:feature_ablation}
\begin{tabular}{|c|c|c|}
\hline
\textbf{Input} & \textbf{General Test} & \textbf{Challenge Test} \\ \hline
RGB \& Flow (1 fps)  & {\bf 84.2} $\pm$ 0.44 & {\bf 59.4} $\pm$ 0.59 \\ \hline
RGB only (1 fps)     & 82.7 $\pm$ 0.47 & 58.0 $\pm$ 0.59 \\ \hline
Flow only (1 fps)    & 64.6 $\pm$ 0.56 & 45.9 $\pm$ 0.53 \\ \hline\hline
RGB \& Flow (Single frame) & 75.6 $\pm$ 0.52 & 51.0 $\pm$ 0.56 \\ \hline
RGB (1 fps) \& Flow (2 fps) & 84.4 $\pm$ 0.44 & 59.6 $\pm$ 0.59 \\ \hline
RGB (2 fps) \& Flow (1 fps) & 84.1 $\pm$ 0.45 & 58.7 $\pm$ 0.59 \\ \hline
RGB (2 fps) \& Flow (2 fps) & 83.7 $\pm$ 0.44 & 59.1 $\pm$ 0.59 \\ \hline
\end{tabular}
\end{table}

\section{Conclusion}
The field of few-shot video action recognition is still quite young.  In
this work, we propose a set of approaches, using different video encoder
architectures and metric-based few-shot methods.  Among the proposed
approaches, a two-stream pooled LSTM-CNN video encoder used with a Prototypical
Network gives the best performance: 84.2\% accuracy on our general test set for
the 5-way 5-shot setup on a few-shot split of the Kinetics 600 dataset.  Given
the inherent computational challenges of processing video,
we find it encouraging that high accuracy can be obtained from a computationally efficient 
few-shot algorithm and a low framerate.  

\section*{Acknowledgment} 
{This work was funded by the U.S. Government.

\newpage
\bibliography{main}{}
\bibliographystyle{plain}

\end{document}